\documentclass{llncs}
\usepackage{array,multirow}
\usepackage{graphicx}
\usepackage{wrapfig}
\usepackage{float}
\usepackage{hyperref}
\usepackage{setspace}
\usepackage{tabularx}
\newcommand\nv{{\bar v}}
\newcommand\tool[1]{{\sc #1}}
\newcommand\implies{\Rightarrow}
\newcommand\bz{{\hat \beta_0}}
\newcommand\bO{{\hat \beta_1}}
\newcommand\new[1]{}
\newcommand\shorter[1]{}
\newcommand\longer[1]{#1}

\begin{document}

\title{The impact of Entropy and Solution Density on selected SAT heuristics}
\author{Dor Cohen and Ofer Strichman}
\institute{Information systems engineering, IE, Technion, Haifa, Israel}
\maketitle

\setlength{\abovecaptionskip}{0pt}
\setlength{\belowcaptionskip}{-14pt}
\setlength{\intextsep}{0.1cm}

\begin{abstract}
In a recent article~\cite{O15}, Oh examined the impact of various key heuristics (e.g., deletion strategy, restart policy, decay factor, database reduction) in competitive SAT solvers.
His key findings are that their expected success depends on whether the input formula is satisfiable or not. To further investigate these findings, we focused on two properties of satisfiable formulas: the \emph{entropy} of the formula, which approximates the freedom we have in assigning the variables, and the \emph{solution density}, which is the number of solutions divided by the search space. We found that both predict better the effect of these heuristics, and that satisfiable formulas with small entropy `behave' similarly to unsatisfiable formulas.
\end{abstract}

\section{Introduction}

In a recent article~\cite{O15}, Oh examined the impact of various key heuristics in competitive SAT solvers.
His key findings are that the average success of those heuristics depends on whether the input formula is satisfiable or not. In particular the effect of the \emph{deletion strategy}, \emph{restart policy}, \emph{decay factor}, and \emph{database reduction} is different, on average, between satisfiable and unsatisfiable formulas. This observation can be used for designing solvers that specialize in one of them, and for designing a hybrid solver that alternates between SAT / UNSAT `modes'. Indeed certain variants of \tool{COMiniSatPS}~\cite{O15} work this way.

We do not see an a priory reason to believe that the SAT/UNSAT divide---corresponding to the distinction between zero or more solutions---explains best the differences in the effect of the various heuristics.\footnote{While proving Unsat and Sat belong to separate complexity classes, there is no known connection of this fact to effectiveness of heuristics.} In this work we investigate further his findings, and
show empirically that there are more refined measures (properties) than the satisfiability of the formula, that predict better the effectiveness of these heuristics. In particular, we checked how it correlates with two measures of satisfiable formulas: the \emph{entropy} of the formula (to be defined below), which approximates the freedom we have in assigning the variables, and the \emph{solution density} (henceforth \emph{density}), which is the number of solutions divided by the search space.
Our experiments show that both are strongly correlated to the effectiveness of the heuristics, but the entropy measure seems to be a better predictor. Generally our findings confirm Oh's observations regarding which heuristic works better with satisfiable formulas. But we also found that satisfiable formulas with small entropy `behave' similarly to unsatisfiable formulas.

\section{Entropy}
\begin{figure}[t]
	\centering
\begin{minipage}{0.49\textwidth}
		\centering
		\includegraphics[scale=0.4]{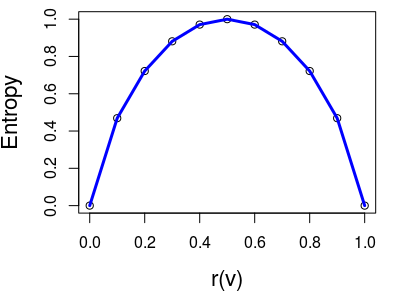}
\end{minipage}
\begin{minipage}{0.49\textwidth}
		\centering
		\includegraphics[height=4cm, width=5.5cm]{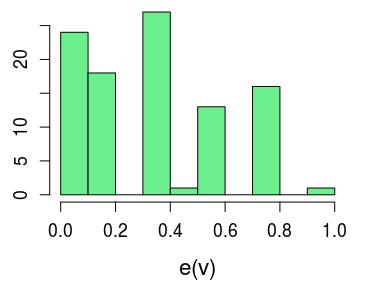}	
\end{minipage}
\caption{(left) Depicting the entropy function (\ref{eq:e}), for a satisfiable formula with 11 solutions. (right) The distribution of $e(v)$ of a formula with 100 variables. \label{fig:ent}}
\end{figure}

Let $\varphi$ be a propositional CNF formula, $var(\varphi)$ its set of variables and $lit(\varphi)$ its set of literals. In the following we will use $v,{\bar v}$ to denote the literals corresponding to a variable $v$ when the distinction between variables and literals is clear from the context. If $\varphi$ is satisfiable, we denote by $r(l)$, for $l \in lit(\varphi)$, the ratio of solutions to $\varphi$ that satisfy $l$. Hence for all $v \in var(\varphi)$, it holds that $r(v) + r(\nv) = 1$.
We now define:

\begin{definition}[variable entropy]
For a satisfiable formula $\varphi$, the \emph{entropy } of a variable $v \in var(\varphi)$ is defined by
\begin{equation} \label{eq:e}
e(v) \doteq -r(v)\log_2 r(v) -r(\nv)\log_2 r(\nv)\;.
\end{equation}
where $0 \cdot \log_2 0$ is taken as being equal to 0.
\end{definition}
This definition is inspired by Shannon's definition of entropy in the context of \emph{information theory}~\cite{S48}. Figure~\ref{fig:ent} (left) depicts (\ref{eq:e}).

Intuitively, entropy reflects how `balanced' a variable is with respect to the solution space of the formula. In particular $e(v) = 0$ when $r(v) = 0$ or $r(v)=1$, which means that $\varphi\implies \nv$ or $\varphi \implies v$, respectively. In other words, $e(v)=0$ implies that $v$ is a \emph{backbone} variable, since its value is implied by the formula. The other extreme is $e(v) = 1$; this happens when $r(v) =r(\nv) = 0.5$, which means that $v$ and $\nv$ appear an equal number of times in the solution space.

\begin{definition}[formula entropy]
The entropy of a satisfiable formula is the average entropy of its variables.
\end{definition}

As an example, Fig.~\ref{fig:ent} (right) is a histogram of $e(v)$ for a particular formula $\varphi$, where for 24 out of the 100 variables $r(v)=0$. 

\subsubsection{Entropy is hard to compute}: Let $\#SAT(\varphi)$ denote the number of solutions a formula $\varphi$ has. Then it is easy to see that
\begin{equation}\label{eq:ev}
  r(v) = \frac{\#(\varphi \land v)}{\#\varphi}  \qquad \mbox{and} \qquad
  r(\bar{v})= 1 - r(v)\;.
\end{equation}
Hence computing $e(v)$ amounts to two calls to a model counter. But since the denominator $\#\varphi$ is fixed for $\varphi$, computing $e(\varphi)$ amounts to  $|var(\varphi)| + 1$ calls to a model counter. Since model counting is a \#P problem, we can only compute this value for small formulas.

\subsubsection{The benchmark set}: Using the model-counter \tool{Cachet}~\cite{SBBKP04}, we computed the precise entropy of 5000 3-SAT random formulas with 100 variables and 400 clauses. These are formulas taken from SAT-lib, in which the number of backbone variables is known. Specifically, there is an equal number of formulas in this set with 10,30,50,70 and 90 backbone variables (i.e., a 1000 formulas of each number of backbone variables), which gave us a near-uniform distribution of entropy among the formulas.

\section{A preliminary: standardized linear regression}\label{sec:prel}
We assume the reader is somewhat familiar with linear regression. It is a standard technique for building a linear model ${\hat y} = \bz + \bO x$, where ${\hat y}$ in our case is a predictor of the number of conflicts, and $x$ is either the entropy or the density of the formula. We will focus on two results of linear regression: the value of $\bO$ and the $p$-value. The latter is computed with respect to a \emph{null hypothesis}, denoted $H_0$, that $\bO=0$, and an alternative hypothesis $H_1$. $H_1$ can be either the complement of $H_0$ ($\bO\neq 0$) or a `one-sided hypothesis', e.g., $H_1: \bO > 0$.
In the former case, $p= 2Pr(Z \leq z \mid H_0)$, where $Z \sim N(0,1)$ and $z = \frac{\bO - 0}{\mbox{std}(\bO)}$. The `0' in the numerator comes from the specific value in $H_0$. In other words, assuming $H_0$ is correct, the $p$-value indicates the probability that a random value from a standard normal distribution $N(0,1)$, is less than $z$, the standardized value of $\bO$. In the latter case $p= Pr(Z \leq z \mid H_0)$.

We list below several important points about the analysis that we applied.
\begin{itemize}
\item {\bf Standardization} of the data: given data points $X \doteq x_1,\ldots,x_n$, their standardization $X' \doteq x_1',\ldots,x_n'$ is defied for $1\leq i\leq n$ by
\shorter{$x_i' = \frac{x_i-{\bar x}}{\sigma}\;,$}
\longer{\[x_i' = \frac{x_i-{\bar x}}{\sigma}\;,\]}
where ${\bar x}$ is the average value of $X$ and $\sigma$ is its standard deviation. Now $X'$ has no units, and hence two standardized sets of data are comparable even if they originated from different types of measures (in our case, entropy and density). All the data in our experiments was standardized.

\item {\bf Bootstrapping}:  Bootstrapping, parameterized by a value $k$, is a well-known technique for improving the precision of various statistics, such as the confidence interval. Technically, bootstrap is applied as follows:
    Given the original $n$ samples, uniformly sample it $n$ times with replacement (i.e., without taking the sampled points out, which implies that the same point can be selected more than once); repeat this process $k$ times. Hence we now have $n\cdot k$ data points. For our experiments we took $k=1000$, which is a rather standard value when using this technique. Hence, we have $5\cdot 10^6$ data points.

\item {\bf Two regression tests}: The entropy and density data consists of pairs of the form $\langle entropy,conflicts[i]\rangle$, and $\langle density,conflicts[i]\rangle$, respectively, where $i \in \{1,2\}$ is the index of the heuristic. Hence the corresponding data is four series of points $(e_1,c_1[i]),\ldots, (e_n,c_n[i])$,
    and  $(d_1,c_1[i]),\ldots, (d_n,c_n[i])$, where $i \in \{1,2\}$. In order to compare the predictive power of entropy, density and Oh's criterion of SAT/UNSAT, we performed two statistical tests (recall that the data is standardized, and hence comparable):
    \begin{itemize}
    \item The $\Delta$ test: A linear regression test over the series $(e_1, c_1[1] - c_1[2]) \ldots$ $(e_n, c_n[1] - c_n[2])$, and the series $(d_1, c_1[1] - c_1[2]) \ldots(d_n, c_n[1] - c_n[2])$.

    \item The $\Delta_\bO$ test: A linear regression test over the series $(e_1, c_1[1]) \ldots(e_n, c_n[1])$ and $(e_1, c_1[2]) \ldots (e_n, c_n[2])$, and similarly for density (i.e., four tests all together). We then checked the significance of $\bO$ for each of these 4 tests (in all such tests the significance was clear). In addition, we checked the hypothesis
        $H_0: \bO[1] - \bO[2] = 0$ for each of the measures. The result of this last test is what we will list in the results table in Appendix~\ref{app:reg}.
    \end{itemize}
    Intuitively, the two models tell us slightly different things: the first tells us whether the gap between the two heuristics is correlated with the measure, and the second tells us whether there is a significant difference in the value of $\bO$ (the slope of the linear model) between the two heuristics. As we will see in the results, the $p$-value obtained by these models can be very different.

\item {\bf Plots}: The plots are based on the original (non-standardized) data. To reduce the clutter (from 5000 points), we rounded all values to 2 decimal points and then \emph{aggregated} them. Aggregation means that points $(x,y_1)\ldots(x,y_n)$ (i.e., $n$ points with an equal $x$\longer{value}) are replaced with a single point ($x,$ $\mbox{avg}(y_1\ldots y_n)$). However the trend-lines in the various plots are depicted according to the \emph{original} data, before rounding and aggregation. The statistical significance of these trend-lines appears in Appendix~\ref{app:reg}.
\end{itemize}

\section{Entropy and density predict hardness }
\begin{figure}[H]
	\centering
	\begin{minipage}{0.5\textwidth}
		\centering
		\includegraphics[scale=0.55]{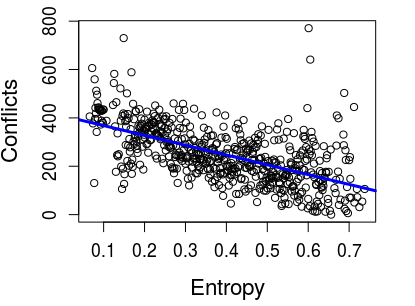}
	\end{minipage}\hfill
	\begin{minipage}{0.5\textwidth}
		\centering
		\includegraphics[scale=0.55]{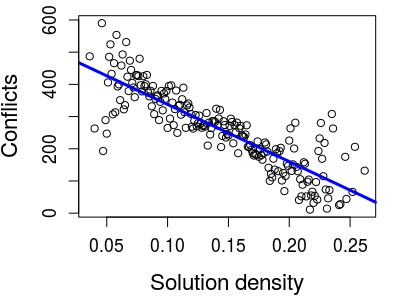}
	\end{minipage}
	\caption{Entropy (left) and density (right) as predictors of the number of conflicts (based on MiniSat-HACK-999ED). It is apparent that higher entropy and higher density imply a smaller number of conflicts. \label{fig:difficulty}}
\end{figure}

We checked the correlation between hardness, as measured by the number of conflicts, and the two measures described above, namely entropy and density. We use the number of conflicts as
a proxy of the run-time, because these are all easy formulas for SAT, and hence the differences in run-time are rather meaningless.
The two plots in Fig.~\ref{fig:difficulty} depict this data based on our experiments with the solver \tool{MiniSat-HACK-999ED}. It is apparent that higher entropy and higher density imply a smaller number of conflicts. A detailed regression analysis appears in Appendix~\ref{app:hardness}, for seven solvers.

We also checked the correlation between the two measures themselves: perhaps formulas with higher entropy also have a higher density (each variable $v$ with high entropy, e.g., $e(v) = 1$, nearly doubles the number of solutions). It turns out that in our benchmarks these two measures are not correlated: the confidence-interval for $\bO$ is [0.144--0.156] with a $p$-value which is practically 0.

\section{Empirical findings}\label{sec:emp}
In this section we describe each of the experiments of Oh~\cite{O15}, and our own version of the experiment based on entropy and density, when applied to the benchmarks mentioned above. We omit the details of one experiment, in which Oh examined the effect of canceling database reduction, the reason being that this heuristic is only activated after 2000 conflicts, and most of our benchmarks are solved before that point.\footnote{Our attempt to use an approximate model-counter with larger formulas failed: the inaccuracies were large enough to make the analysis show results that are senseless.}
Raw data as well as charts and regression analysis of our full set of experiments can be found online in~\cite{CS16}.

\begin{figure}[t]
		\centering
		\includegraphics[scale=0.5]{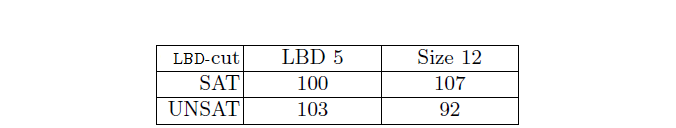}
\begin{minipage}{0.49\textwidth}
		\centering
\includegraphics[scale=0.5]{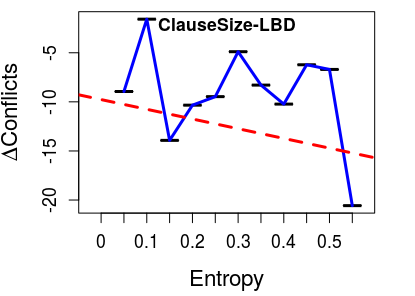}			
\end{minipage}
\begin{minipage}{0.49\textwidth}
		\centering
\includegraphics[scale=0.5]{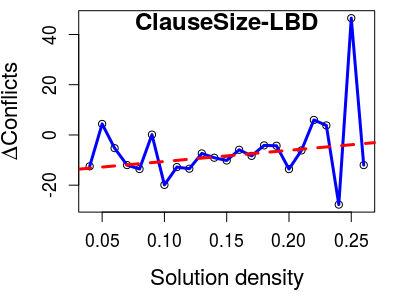}	
\end{minipage}		
\caption{The effect of the deletion criterion. The results of~\cite{O15} appear in the table at the top of the figure (the numbers indicate the solved instances). It shows that for SAT instances keeping everything with clause size 12 is better than keeping everything with LBD 5, whereas the result is opposite for the UNSAT instances. Our own experiments (bottom left) show that within SAT instances, the clause size criterion becomes better with higher entropy, but not with higher density. Note that the y-axis corresponds to the difference in the \# of conflicts. On average on these instances the \# of conflicts across both methods was $\approx 290$. \label{fig:T2}}
\end{figure}

\vspace{0.1 cm}
{\bf 1. Deletion strategy}:  Different solvers use different criteria for selecting the learned clauses for deletion. It was shown in~\cite{O15} that for SAT instances learned clauses with low Literal Block Distance (LBD)~\cite{AS09} value can help, whereas others have no apparent effect. In one of the experiments, whose results are copied here at the top part of Fig.~\ref{fig:T2}, Oh compared the criterion of `core \emph{LBD-cut}'\footnote{An LBD-cut is the lowest value of LBD a learned clause had so far, assuming this value is recalculated periodically.} 5 and clause size 12. In other words, either save (i.e., do not delete) clauses with an LBD-cut of 5 and lower, or clauses with size 12 or lower.
It shows that for UNSAT instances the former is better, whereas the opposite conclusion is reached for the SAT instances. The results of our own experiments are depicted at the bottom of the figure. They show that the latter is indeed slightly better with our benchmarks (all satisfiable, recall). But what is more important, is that the difference becomes smaller with lower entropy---hence the decline of the trend-line (recall that the trend-lines are based on the raw data, whereas the diagram itself is computed after rounding and aggregation to improve visibility). Hence it is evident that formulas with small entropy `behave' more similar to unsat formulas. The ascending trend-line in the right figure shows, surprisingly, an opposite effect of density.

\begin{figure}[t]
	\centering	
    \includegraphics[scale=0.5]{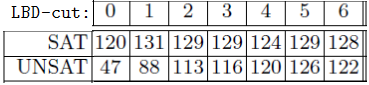}

    \begin{minipage}{0.49\textwidth}
	\centering		
    \includegraphics[scale=0.6]{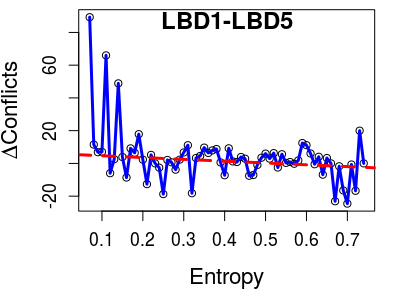}
    \end{minipage}
    \begin{minipage}{0.49\textwidth}
    \includegraphics[scale=0.6]{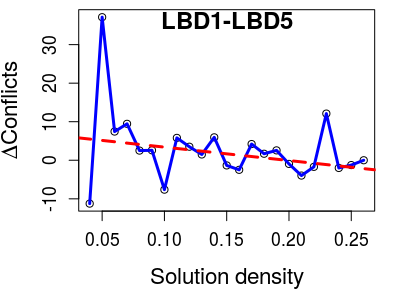}
    \end{minipage}
    \caption{The results of~\cite{O15} (top) show that unsat formulas are solved faster with high LBD-cut. Our results (bottom) show that low-entropy and low-density formulas behave more similarly to unsat formulas. }\label{fig:lbdcut}
\end{figure}

\vspace{0.1 cm}
{\bf 2. Deletion with different LBD-cut value} Related to the previous heuristic, in~\cite{O15} it was found that deletion based on larger LBD-cut values, up to a point, improve the performance of the solver with unsat formulas, but not with SAT ones. Fig.~\ref{fig:lbdcut} (top) is an excerpt from his results for various LBD-cut values. We repeated his experiment with LBD-cut 1 and LBD-cut 5. The plots show that lower values of entropy and (independently) lower values of density yield a bigger advantage to LBD-cut 5, which again demonstrates that satisfiable formulas with these values `behave' similarly to unsat formulas.

\vspace{0.1 cm}
{\bf 3. Restarts policy}: The Luby restart strategy~\cite{LSZ93} is based on a fixed sequence of time intervals, whereas the \tool{Glucose} restarts are more rapid and dynamic. It initiates a restart when the solver identifies that learned clauses have higher LBD than average. According to the competitions' results this is generally better in unsat instances. Oh confirmed the hypothesis that this is related to the restart strategy: indeed his results show that for satisfiable instances Luby restart is better.

Our own results can be seen in Fig.~\ref{fig:restarts} and in Appendix~\ref{app:reg}. The fact that the gap in the number of conflicts between Luby and Glucose-style restarts is negative, implies that the former is generally better, which is consistent with Oh's results for satisfiable formulas.
Observe that the trend-line slightly declines with entropy ($\bO = -15$), which implies that
Glucose restarts are slightly better with low entropy. So again we observe that low entropy formulas `behave' more similar to UNSAT formulas than those that have high entropy. The table in Appendix~\ref{app:reg} shows that this result has a relatively high $p$-value.
We speculate that with high-entropy instances, the solver hits more branches that can be extended to a solution, hence Glucose's rapid restarts can be detrimental. Density seems to have an opposite effect, although again only with low statistical confidence.

\begin{figure}[t]
	\centering	
    \includegraphics[scale=0.5]{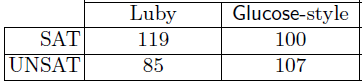}

\begin{minipage}{0.49\textwidth}
	\centering		
    \includegraphics[scale=0.6]{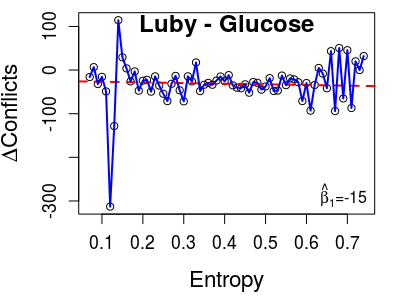}
	\end{minipage}
\begin{minipage}{0.49\textwidth}
    \includegraphics[scale=0.6]{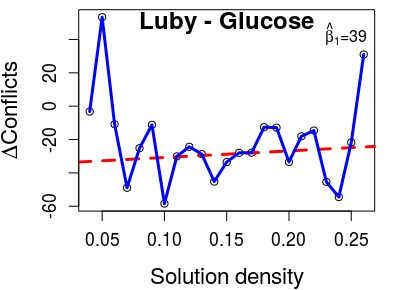}
\end{minipage}
\caption{The effect of the restart strategy, comparing Luby and Glucose-style restarts. The results of~\cite{O15} (top) show that the Glucose strategy (rapid restarts) has an advantage in unsat formulas.
Our results (bottom) show that the same phenomenon is apparent in formulas with low entropy. Indeed observe that the number of conflicts with Glucose becomes smaller than it is with Luby (hence the negative gap), in satisfiable formulas with low entropy. \label{fig:restarts}}
\end{figure}

\begin{figure}[h!]
	\centering	
\includegraphics[scale=0.6]{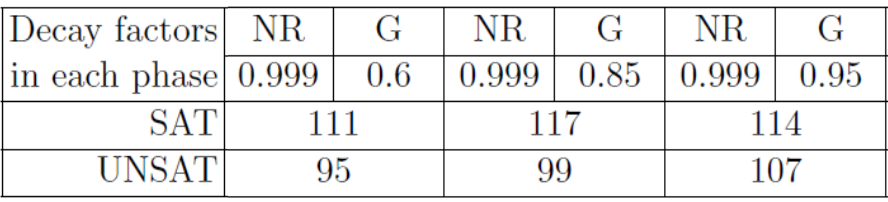}\\

\begin{minipage}{0.49\textwidth}
		\centering		
\includegraphics[scale=0.6]{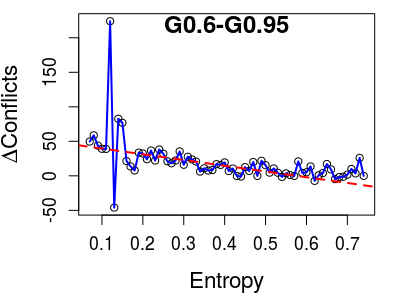}	
\end{minipage}
\begin{minipage}{0.49\textwidth}
		\centering
\includegraphics[scale=0.6]{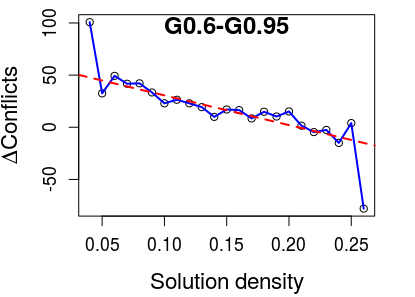}	
\end{minipage}
\longer{\caption{The effect of variable decay: the results of~\cite{O15} (top) generally show that unsat formulas are better solved with a high \emph{decay factor}. The restart policy in his solver is hybrid: it alternates between a `no-restart' (NR) phase and a `Glucose' (G) phase.  The `NR' and `G' columns hold the decay factor during these phases.
The plots at the bottom show the gap in the number of conflicts between $G=0.6$ and $G=0.95$. It shows that with low entropy, strong decay (i.e., $G=0.6$) is worse, similar to the effect that it has on unsat formulas. With low density (right) a similar effect is visible.
}}
\shorter{
\caption{The effect of variable decay: the results of~\cite{O15} (top), and our own (bottom).  Our results show that with low entropy, strong decay (i.e., $G=0.6$) is worse, similar to the effect that it has on unsat formulas. With low density (right) a similar effect is visible.
}}
\label{fig:decay}
\end{figure}

\vspace{0.1 cm}
{\bf 4. The variable decay factor}: The well-known VSIDS branching heuristic is based on an activity score of literals, which decay over time, hence giving higher priority to literals that appear in recently-learned clauses. In the solver \tool{MiniSat\_HACK\_999ED}, there is a different decay factor for each of the two \emph{restart phases}: this solver alternates between a Glucose-style (G) restart policy phase and a no-restart (NR) phase (these two phases correspond to good heuristics for SAT and UNSAT formulas, respectively).
In~\cite{O15} Oh compares different decay factors for each of these restart phases\longer{, on top of \tool{MiniSat\_HACK\_999ED}}. His results show that for UNSAT instances slower decay gives better performance, while for SAT instances it is unclear. His results appear at the top of Fig.~\ref{fig:decay}. We experimented with the two extreme decay factors in that table: 0.95 and 0.6. Note that since our benchmarks are relatively easy, the solver never reaches the NR phase. The plot at the bottom of the figure shows the gap in the number of conflicts between these two values. A higher value means that with strong decay (0.6) the results are worse. We can see that the results are worse with strong decay when the entropy is low, which demonstrates again that the effect of the variable decay factor is similar for unsat formulas and satisfiable formulas with low entropy. A similar phenomenon happens with small density.

\subsubsection{Conclusions}:
We defined the \emph{entropy} property  of satisfiable formulas, and used it, together with solution density, to further investigate the results achieved by Oh in~\cite{O15}. We showed that both are strongly correlated with the difficulty of solving the formula (as measured by the number of conflicts). Furthermore, we showed that they predict better the effect of various SAT heuristics than Oh's sat/unsat divide, and that satisfiable formulas with small entropy `behave' similarly to unsatisfiable formulas. Since both measures are hard to compute we do not expect these results to be applied directly (e.g., in a portfolio), but perhaps future research will find ways to cheaply approximate them. For example, a high backbone count (variables with a value at decision level 0) may be correlated to low entropy, because such variables contribute 0 to the formula's entropy.

\longer{
\subsubsection{Acknowledgement}
We thank Dr. David Azriel for his guidance regarding statistical techniques.
}

\bibliographystyle{abbrv}
\bibliography{main}
\appendix
\section{Predicting hardness: a regression analysis}\label{app:hardness}
 Denote by $\bO^E$ and $\bO^S$ the $\bO$-value of the linear models for entropy vs. conflicts and density vs. conflicts, respectively. The table below shows strong correlation between both measures to the number of conflicts (the $p$-value in both cases, for all engines, is practically 0). The Last two columns show the gap $\bO^E - \bO^S$ and the corresponding $p$-value for $H_0: \bO^E-\bO^S=0, H_1: \bO^E-\bO^S\neq 0$, when measured across the $k=1000$ iterations of the bootstrap method that was described in Sec.~\ref{sec:prel}. For engines with high $p$-value we cannot reject $H_0$ with confidence.

\begin{table}
\centering
\begin{tabular}{|c|c|c||c|c|}  \hline
Solver	& $\bO^E$	& $\bO^S$	& $\bO^E-\bO^S$	& $p$-value    \\ \hline
MiniSat-HACK-999ED & (-84.29, -72.58 )	& (-84.93, -73.56 )	& ( 5.37, 16.96 )	& 0.716\\ \hline
MiniSat-HACK-999ED &&&&\\
(modified to luby)	& (-86.31, -75.36 )	& (-82.97, -72.64 )	& (-7.51,  1.44 )	& 0.200\\ \hline
MiniSat-HACK-999ED &&&&\\
(modified for 2 phases)	&  (-72.84, -63.61 )	& (-72.31, -62.91 )& (-4.80,  3.57 )	& 0.738\\ \hline
SWDiA5BY	& (-91.61, -79.17 )	& (-90.97, -78.77 )	& (-5.95,  4.92 ) & 0.84\\ \hline
COMiniSatPS	& (-74.68, -64.58 )	& (-75.41, -65.43 )	& (-3.79,  5.37 ) & 0.76\\ \hline
lingeling-ayv	& (-76.19, -66.61 )	& (-71.70, -61.76 )	& (-8.99, -0.35 )	& 0.029 \\ \hline

Glucose	& (-91.24, -79.34 )	& (-90.56, -78.88 )	& (-6.00,  4.85 )
& 0.845\\ \hline
\end{tabular}
\caption{For each solver, we list the 95\% confidence interval of its $\bO^E$ (entropy) and $\bO^S$ (solutions). For all engines the corresponding $p$-value is practically 0 (i.e.,. $\leq 10^{-100}$). The last two columns refer to the gap between these measures. }\label{tab:hardness}
\end{table}

\section{Regression-tests results}\label{app:reg}
The table below lists the confidence interval and corresponding $p$-value, for the two regression tests $\Delta$ and $\Delta_\bO$ (in the latter we also list the results for $\bz$) that were explained in Sec.~\ref{sec:prel}, and the four experiments described in Sec.~\ref{sec:emp}. $H_1$ is one-sided.
\begin{table}
\centering
    \begin{tabular}{|l|l||c|c||c|c|c|c|}  \hline
    Exp. & Measure & Conf. interval  & $p$-val &  Conf. interval & $p$-val &  Conf. interval & $p$-val \\
    &&($\Delta$)&&($\Delta_\bO$)&&($\Delta_\bz$)& \\
    \hline \hline

    1 & Entropy & (-2.76,  2.64 )   & 0.48    & (-2.75, 2.46)  &  0.05 & (-12.06, -6.60) & 0 \\ \hline
      & Density & (-0.81,  4.35 )   & 0.09    & (-0.83, 4.43)  &  0.39  & (-12.06, -6.64) & 0\\ \hline

    2 & Entropy & (-3.72,  0.25 )   & 0.04     & (-3.78,  0.25 ) & 0.39 & (0.48, 4.61 ) & 0.01 \\ \hline
      & Density & (-3.40,  0.59 )   & 0.09     & (-3.34,  0.69 ) & 0.47 & (0.47, 4.56 )& 0.01 \\ \hline

    3 & Entropy & (-8.31,  3.52 )   & 0.22    & (-8.01,  3.67 )    &  0.001  & (-36.12, -23.78 ) & 0\\ \hline
      & Density & (-4.41,  7.36 )   & 0.30    & (-4.34,  7.50 )     &  0.05 & (-35.99, -23.90 ) & 0\\ \hline

    4 & Entropy & (-15.1, -10.6 ) & 0  & (-15.1, -10.7 )   &  0.125 & (20.99, 25.44 ) & 0 \\ \hline
      & Density & (-3.92,  0.60  )  & 0  & (-13.60, -8.86 )    & 0.475  & (20.96, 25.47 ) & 0\\ \hline

    \end{tabular}
    \caption{Regression-tests results for the four experiments in Sec.~\ref{sec:emp}. $p$-value $\leq 10^{-10}$ are rounded to 0.}
\end{table}
\end{document}